\documentclass[preprint,11pt]{elsarticle}

\usepackage{wrapfig}

\usepackage{amssymb}
\usepackage{amsfonts}
\usepackage{nicefrac}

\usepackage{times}
\usepackage{float}
\usepackage{amsmath}
\usepackage{changepage}
\usepackage{graphicx}
\usepackage{caption}
\usepackage{relsize}
\usepackage{times}
\usepackage{subfigure}
\usepackage{natbib}
\usepackage{hyperref}
\usepackage{mathtools}

\newcommand{\cev}[1]{\reflectbox{\ensuremath{\vec{\reflectbox{\ensuremath{#1}}}}}}

\DeclarePairedDelimiter\abs{\lvert}{\rvert}%
\makeatletter
\let\oldabs\abs
\def\abs{\@ifstar{\oldabs}{\oldabs*}}
\let\oldnorm\norm
\def\norm{\@ifstar{\oldnorm}{\oldnorm*}}
\makeatother

\journal{Automation in Construction}

\makeatletter
\def\ps@pprintTitle{%
 \let\@oddhead\@empty
 \let\@evenhead\@empty
 \def\@oddfoot{}%
 \let\@evenfoot\@oddfoot}
\makeatother
\begin{document}

\begin{frontmatter}



\title{Automatically Learning Construction Injury Precursors from Text \\ {\small\textit{Accepted for publication in Automation in Construction}}}

\author[label1]{Henrietta Baker}
\author[label2]{Matthew R. Hallowell}
\author[label3,label4]{Antoine J.-P. Tixier}
\address[label1]{University of Edinburgh, UK}
\address[label2]{University of Colorado at Boulder, USA}
\address[label3]{\'{E}cole Polytechnique, France}
\address[label4]{corresponding author}

\begin{abstract}

\textbf{Highlights}
\begin{itemize}
    \item We propose several methods to automatically learn injury precursors from raw construction accident reports,
    \item precursors are learned by multiple supervised models as a by-product of training them at predicting safety outcomes,
    \item we experiment with two deep learning models (CNN and HAN), and a more traditional machine learning approach (TF-IDF + SVM),
    \item the proposed methods can also be used to visualize and understand the models' predictions.
\end{itemize}

In light of the increasing availability of digitally recorded safety reports in the construction industry, it is important to develop methods to exploit these data to improve our understanding of safety incidents and ability to learn from them.
In this study, we compare several approaches to automatically learn injury precursors from raw construction accident reports. 
More precisely, we experiment with two state-of-the-art deep learning architectures for Natural Language Processing (NLP), Convolutional Neural Networks (CNN) and Hierarchical Attention Networks (HAN), and with the established Term Frequency - Inverse Document Frequency representation (TF-IDF) + Support Vector Machine (SVM) approach.
For each model, we provide a method to identify (after training) the textual patterns that are, on average, the most predictive of each safety outcome. We show that among those pieces of text, valid injury precursors can be found.
The proposed methods can also be used by the user to visualize and understand the models' predictions.
\end{abstract}

\begin{keyword}
construction safety \sep knowledge extraction \sep deep learning \sep natural language processing \sep text mining \sep artificial intelligence


\end{keyword}

\end{frontmatter}


\section{Introduction}
More data than ever are digitally available. An analysis by the International Data Group \cite{reinsel2018dataage} predicts that the amount of digital data will grow from 33 billion Terabytes (TB) in 2018 to 175 billion TB by 2025. Additionally, much of these data are in unstructured format including audio, video and free-text. This proportion is widely quoted as 80\% \cite{grimes80unstructured}. Information overload, where the volume of data produced has outgrown their processing and analysis capacity \cite{woods2002can}, is a growing concern for many industries, especially in the case of free-text data which traditionally relies on human oversight to extract actionable information. Henke et al. \cite{henke2016age} estimate that 76\% of work activities require natural language understanding; therefore, developing automated methods to efficiently process natural language texts is essential.

The construction industry is not immune to this upward trend in data volume and the desire to use it. One area in which there is increasing interest in exploiting written text is that of construction safety, specifically digital injury reports. Learning from incidents is acknowledged to be a key factor in preventing future injuries (e.g. \cite{lukic2012framework, sanne2008incident}) and exploiting data collected about safety incidents on projects is essential to this process. Improving safety is necessary in an industry where the rate of occupational fatality is around 3 times the national average for main industries in both the USA and Europe \cite{william2014fatality}.  

Recently, attempts have been made to introduce Natural Language Processing (NLP) methods to free-text data in the construction industry for better retrieval and analysis of documents. In this study, we compare several approaches to automatically learn injury precursors from raw construction accident reports. The contributions of this paper are listed below.

\begin{itemize}

\item We propose an approach to automatically extract valid accident precursors from a dataset of raw construction injury reports. To the best of our knowledge, this has not been done in construction before. Such information is highly valuable, as it can be used to better understand, predict, and prevent injury occurrence.
\item We experiment with two state-of-the-art deep learning architectures for NLP: Convolutional Neural Networks (CNN) and Hierarchical Attention Networks (HAN). To the best of our knowledge, this study is the first published application of supervised deep learning architectures to construction text. We also experiment with the established Term Frequency - Inverse Document Frequency representation (TF-IDF) + Support Vector Machine (SVM) pipeline.
\item For each model, we propose a method to identify, after training, the textual patterns that have been learned to be, on average, the most predictive of each safety outcome. We verify that valid precursors can be found within these text fragments and make several suggestions to improve the results.
\item The proposed methods can also be used to visualize and understand the models' predictions.
\item Predictive skill is high for all models. Interestingly, we observe that the simple TF-IDF + SVM approach is on par with (or outperforms) deep learning most of the time.
\end{itemize}

To demonstrate the aforelisted contributions, this paper is structured as follows. First, we provide some background about the use of deep learning in NLP, and review the relevant construction literature. We then present the models we compared, introduce the dataset and the preprocessing employed, and detail our experimental setup. We then report and interpret the results, and propose, for each model, a method to extract precursors from text. We finally give recommendations for further development.

\subsection{Background: deep learning in NLP}\label{sub:background}
NLP, also known as computational linguistics, is a rapidly developing field dealing with the computer analysis of both written and spoken human language. It is acknowledged to be an interdisciplinary field, using concepts from linguistics as well as computer science, statistics, and machine learning in general. As well as applications in speech recognition and machine translation, NLP has gained interest in text retrieval and automated content analysis.

Goodfellow \cite{goodfellow2016deep} defines deep learning as machine learning methods which \textit{``allow computers to learn complicated concepts by building them out of simpler ones''}. If represented graphically, these models have many layers, the number of which is referred to as depth. Hence if a model has multiple layers, it is referred to as \textit{deep}. Deep learning methods typically rely on large quantities of data to train their parameters. For that reason, their rise to power coincides with the global increase in data availability and computational power.

Neural networks are the most common collection of deep learning architectures and often the terms are used interchangeably. While they were initially developed as early as 1940s, these simplistic early networks have undergone radical developments and increases in levels of sophistication, achieving record pattern recognition levels since the 1990s \cite{schmidhuber2015deep,lecun1998gradient}.

It is not until recently that, following the advent of distributed word representations \cite{bengio2003neural,mikolov2013distributed,mikolov2013efficient}, deep learning architectures have been developed for NLP tasks such as natural language understanding and machine translation (with great success) \cite{kim2014convolutional,luong2015effective}. 
Next, we give a brief evolution of text representation, from vector space to word embeddings.

\noindent \textbf{Bag-of-Words (BoW)}. Transforming unstructured free-text data into a structured representation is a key preliminary task in many NLP applications. While early researchers focused on writing lexical rules which computers could follow, this was found in most reports to be unwieldy due to word ambiguity and grammatical complexity, giving rise to the popularity of empirical language models in the late 1980s \cite{hirschberg2015advances,katz1987estimation}.
Since then, such empirical models, based on the \textit{Bag-of-Words} (BoW) representation (also known as the \textit{vector space} representation), have occupied the limelight owing to the notable results found when trained on large quantities of data. 

With BoW, a given document is represented as a vocabulary-size vector that has zeroes everywhere except for the dimensions corresponding to the words in the document.
The vocabulary is made of all the unique words in the preprocessed training set. Depending on preprocessing, words may include phrases, punctuation marks, numbers, codes, etc. For this reason, the term \textit{token} is often used in lieu of \textit{word}.

BoW ignores word similarity and word order. For example, ``hammer fell on worker'' and ``worker fell on hammer'' have the same representation, and ``hammer'' and  ``tool'' are not considered more similar than ``hammer'' and ``worker'', as all dimensions of the vector space are orthogonal. This restricts the semantic meaning which can be gained from such representations. To capture word order locally, combinations of tokens (i.e., phrases), formally known as $n$-grams, may be used instead of single tokens. But doing so makes the vector space become so large and sparse that it makes it hard to fit any model, a problem colloquially known as the \textit{curse of dimensionality}. In practice, it is rarely possible to use $n$-grams of order greater than 4 or 5.

Finally, some syntactic information may be captured by creating different dimensions for the different \textit{part-of-speech} tags of a given unigram (noun, proper noun, adjective, verb, etc.), but this has the same adverse effects on the dimensionality of the space as that previously mentioned.

\noindent \textbf{Word embeddings}.
New ways of representing textual data are based on \textit{embeddings} \cite{bengio2003neural,mikolov2013distributed,mikolov2013efficient}, also known as \textit{word vectors} or \textit{distributed word representations}. With word embeddings, each word in the vocabulary is represented as a small, dense vector, in a space of shared concepts. To derive a representation for a document, the vectors of its words are combined, either simply through averaging or concatenation, or through more sophisticated operations (neural networks). One should note that character or subword embeddings are sometimes used \cite{zhang2015character,bojanowski2017enriching}, with the main benefit of providing robustness to out-of-vocabulary words and typographical errors. However, the word is the most common granularity level.

Unlike the long and sparse BoW vectors, word vectors are short (typically 100-500 entries), dense, and real-valued. The dimensions of the embedding space are shared latent features, so that after training, meaningful semantic and syntactic similarities (see Table \ref{table:wv_san}), and other linguistic regularities, are captured. For instance, \cite{tixier2016word} applied the unsupervised \texttt{word2vec} \cite{mikolov2013distributed} model to a large corpus of construction-related text. In the final embedding space, a constant linear translation was found to link body parts (tendon, brain) to sustained injuries (tendonitis, concussion), and another one to link tools and equipment (grinder, chisel) to the corresponding material (metal, wood).

Deep learning architectures are fed the sequence of word vectors of the input document and pass them through their layers. Each layer computes a higher-level, more abstract representation of the input text by performing operations (e.g. convolutional, recurrent) on top of the output of the previous layer, until a single vector representing the entire input document is obtained.

Then, depending on the task, one may add a few specific final layers (e.g. dense, sigmoid, softmax for regression or classification), a decoder (sequence-to-sequence setting for translation or summarization), or combine two encoders via a meta-architecture (e.g. siamese or triplet configuration for textual similarity \cite{shang2019energy}). The specific architectures used in this paper are presented in sections \ref{CNN_sec} and \ref{HAN_sec}.

The word vectors are not necessarily initialized at random, like the other parameters of the network. It is actually advantageous to pre-train them in an unsupervised way. Then, word vectors can either be fine-tuned or kept frozen during training. When pre-training is conducted on an external, typically large corpus of unannotated raw text, the approach is known as \textit{transfer learning}. To this purpose, unsupervised, shallow models such as \texttt{word2vec} \cite{mikolov2013distributed} or \texttt{GloVe} \cite{pennington2014glove} can be applied to big corpora like entire Wikipedia dumps or parts of the Internet\footnote{e.g., \url{https://code.google.com/archive/p/word2vec/} (under section ``pre-trained word and phrase vectors''), \url{https://nlp.stanford.edu/projects/glove/}}.
\texttt{ELMo} \cite{peters2018deep} and \texttt{BERT} \cite{devlin2018bert} have also made great strides recently, by showing that it was possible to transfer not only the word vectors but the entire model. After pre-training, the model (or simply its internal representations in the case of \texttt{ELMo}) are used in a supervised way to solve some downstream task, for example, sentiment analysis or named entity recognition.
\texttt{ELMo} and \texttt{BERT} have brought great improvement to many natural language understanding tasks.

\subsection{Background: NLP in Construction}
Global interest has grown in applying NLP for comprehension and analysis of construction documents. Examples can be found for both retrieval and classification in Table \ref{table:NLP_in_construction}. However, nearly all examples found use the BoW representation (with TF-IDF weighting), losing the semantic relationships between words and ignoring word order.

The literature found showcase a number of different natural language tasks performed in the construction sector, including classification and retrieval of documents. \cite{caldas2003automating, goh2017construction, zhang2019construction} compare machine learning classifiers using TF-IDF inputs and all find that Support Vector Machines (SVM) result in the highest accuracy. In other classification tasks, \cite{chokor2016analyzing, marzouk2019text} elected to cluster the TF-IDF vectors. For two document retrieval tasks, the researchers used vector similarity to identify the most relevant reports \cite{yu2013content, zou2017retrieving}.

Some studies attempted to adjust for the shortcomings of the BoW+TF-IDF representation. \cite{zou2017retrieving} and \cite{kim2019accident} attempted to recapture some semantic relations by implementing thesaurus relations into their BoW vectors; however, this required the use of construction specific dictionaries to supplement thesaurus definitions from general lexicons due to the specificity of construction language. \cite{williams2014predicting} incorporated bigrams into their text representation in order to capture some of the local word order; however, they found that higher level word groupings were unable to significantly increase the accuracy of the predictions. Finally,  \cite{tixier2016word} used a Wasserstein distance in the word embedding space for injury report retrieval and classification (with the $k$-nearest neighbors algorithm, for classification).

Meanwhile, \cite{tixier2016application} and  \cite{tixier2017construction} extracted 81 fundamental attributes (or precursors) from injury reports using a tool based on an entirely hand-written lexicon and set of rules \cite{tixier2016automated}. This allowed them respectively to predict safety outcomes with good skill, and to identify interesting combinations of attributes, coined as ``safety clashes''. However, the development of the tool was resource intensive, both in terms of time and human-input requirement.

\begin{table}[h]

\centering
\renewcommand{\arraystretch}{1.5}
\scalebox{0.7}{
\begin{tabular}{|p{3.2cm}|p{5.75cm}|p{3cm}|p{4.5cm}| }
\hline
  Reference   &   Task   &  Representation  & Analysis algorithm(s) \\
     \hline
 \cite{caldas2003automating} Caldas03  & Classification of management documents  & BoW+TF-IDF  & NB, k-NN, Rocchio, SVM \\
 \cite{chokor2016analyzing} Chokor16 & Classification of accident reports & BoW+TF-IDF &  Unsupervised clustering  \\
 \cite{goh2017construction} Goh15 &  Classification of accident reports  & BoW+TF-IDF & NB, k-NN, RF, LR, SVM \\
 {\footnotesize \cite{kim2019accident,moon2018analysis} Kim19, Moon19} & Retrieval of accident reports  & BoW+TF-IDF* & Rule-based, CRF \\
 \cite{marzouk2019text} Marzouk19  & Classification of contractual documents & BoW+TF-IDF & Clustering \\
{\footnotesize \cite{tixier2016automated} + \cite{tixier2016application} Tixier16-16}  & Prediction of safety outcomes  & 81 attributes & RF and Boosting \\
 \cite{tixier2016word} Tixier16  & Classification/retrieval of accident reports & word vectors & Word Mover's Distance, k-NN \\
 {\footnotesize \cite{tixier2016automated} + \cite{tixier2017construction} Tixier16-17}  & Attribute clustering & 81 attributes & Community detection, hierarchical clustering \\
 \cite{williams2014predicting} Williams14  &  Prediction of cost overruns & BoW+TF-IDF**  & Riddor, K-Star, RBF neural nets \\
 \cite{yu2013content} Yu13  & Retrieval of accident reports  &  BoW+TF-IDF  &  Vector similarity \\
 \cite{zhang2019construction} Zhang19 &  Classification of accident reports & BoW+TF-IDF  & NB, k-NN, RF, LR, SVM \\
 \cite{zou2017retrieving} Zou17 & Retrieval of accident reports  &  BoW* & Vector similarity \\
 \hline
\end{tabular}
}
\captionsetup{size=footnotesize}
\caption{NLP construction literature. Acronyms: Naive Bayes (NB), k-Nearest Neighbor (k-NN), Random Forest (RF), Linear Regression (LR), Support Vector Machine (SVM), Conditional Random Fields (CRF).
*with word2vec and thesaurus implementation, **with bigrams. \label{table:NLP_in_construction}}
\end{table}

These works demonstrate the potential of NLP in the construction domain. None of them, however, experiment with deep learning methods, despite the fact that, as highlighted in the previous section, neural representations have much more expressive power than BoW-based ones. 

There has been limited experimentation with deep learning methods for text data in related subject areas, such as \cite{Cheng2018Bridge} who apply recurrent neural networks with LSTM units to perform named entity recognition on bridge inspection reports. This is not construction text; however, future research in these related civil engineering fields could yield some insights. 

Next, we introduce the dataset used in our study,  explain how we performed preprocessing, present the models we experimented with, report and interpret the results, and explain, for each model, how to extract injury precursors from raw text.

\section{Models} 

\noindent In what follows, we present the three models we experimented with. The two deep learning models are structurally different, being respectively feedforward (convolutional) and recurrent. 

\subsection{Convolutional Neural Network (CNN)} \label{CNN_sec}
\begin{figure}[h]
\centering
\includegraphics[width=0.75\textwidth]{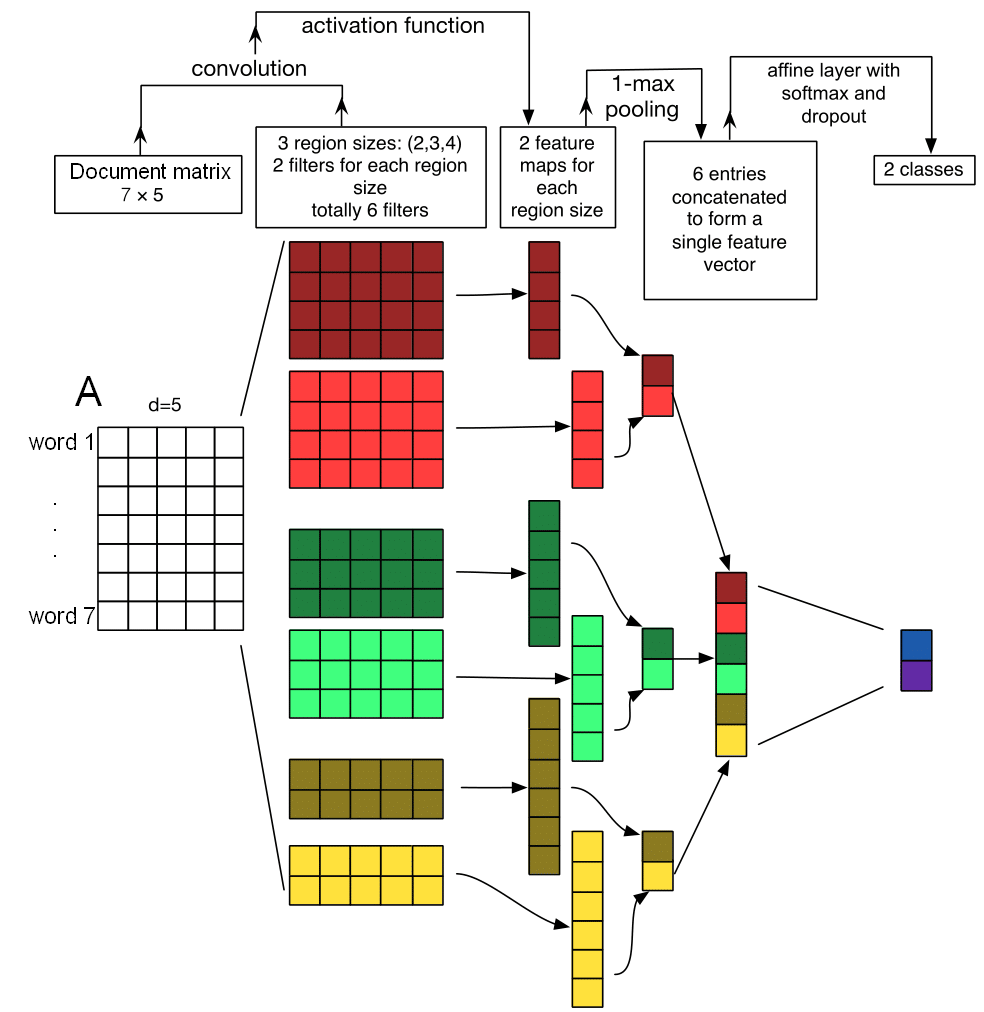}
\captionsetup{size=footnotesize}
\caption{\label{fig:cnn} Toy example of CNN architecture for document classification, with an input document of size $s=7$, word vectors of dimensionality $d=5$, 3 branches corresponding to filters of sizes $h=\big\{2,3,4\big\}$, $n_f=2$ filters per branch, $1$-max global pooling, and 2 categories. Adapted with permission from \cite{zhang2015sensitivity}.}
\end{figure}

The CNN architecture, shown in Fig. \ref{fig:cnn}, was initially developed for inputs with two spatial dimensions in Computer Vision \cite{lecun1998gradient}, and later brought to the NLP domain \cite{kim2014convolutional,johnson2014effective}, where the inputs are spatially unidimensional.

\noindent\textbf{Input}. The input document is represented as a real matrix $A \in \mathbb{R}^{s \times d}$, where $s$ is the document length in number of words (spatial dimension), and $d$ is the dimension of the word embedding space (depth dimension). Following common practice, we fixed $s$ at the dataset level and used truncation and zero-padding respectively for longer and shorter documents. The value of $s$ was chosen to be 200 after inspecting the distribution of number of words per document.

\noindent\textbf{Convolution layer}.
Each instantiation of a one-dimensional window of size $h$ slided over the input document is multiplied elementwise with $n_f$ parameter matrices called \textit{filters}, initialized randomly and learned during training. Just like the window, the filters have only one spatial dimension, but they extend fully through the depth dimension. The instantiations of the window over the input are called \textit{regions} or \textit{receptive fields}. When sliding the window one word at a time, there are $s-h+1$ such regions. The output of the convolution layer for a given filter is thus a vector $o \in \mathbb{R}^{s-h+1}$.
Then, a nonlinear activation function $f$ is applied elementwise to $o$, returning the \textit{feature map} $c \in \mathbb{R}^{s-h+1}$ associated with the filter. In our experiments, we used the \texttt{ReLU} nonlinearity as our activation function ($x \mapsto \mathrm{max}(0,x)$), which is a standard choice.

\noindent\textbf{Pooling layer}. Following \cite{zhang2015sensitivity}, we used \textit{global $k$-max pooling} with $k=1$ to extract the greatest value from each feature map, as shown in Fig. \ref{fig:cnn}. These values were concatenated into a final vector. Global 1-max pooling makes the model focus on the most salient feature from each feature map. Also, one should note that the pooling operation in general makes the model lose track of the position of the feature in the input document. The assumption is that it does not matter, and that what matters is only whether the feature is present or absent. For instance, under this assumption, whether ``congested workspace'' appears at the beginning or at the end of a report does not matter, as long as it appears.

Per the recommendations of \cite{zhang2015sensitivity}, we implemented 3 branches featuring respectively filters of size 2, 3, and 4. Moreover, we used 100 filters per branch for all outcomes except \texttt{incident\_type}, for which we used 300 filters. Indeed, the much larger size of the training set for \texttt{incident\_type} (82,474 reports compared to $<$ 17,500 for the other outcomes) allowed using a model with more parameters without risking to overfit the data. Using many filters of different sizes gave the models the ability to learn different, complementary features for each region. The branch outputs were concatenated after pooling, as shown in Fig. \ref{fig:cnn}. 

\subsection{Hierarchical Attention Network (HAN)} \label{HAN_sec}

Recently, hierarchical architectures have set new state-of-the-art on many NLP tasks. Such architectures build on the assumption that the representation of the input text should be learned in a bottom-up fashion by using a different encoder at each granularity level (words, sentences, paragraphs), where each encoder takes as input the output of the preceding encoder. 

\begin{figure}[h]
\centering
\includegraphics[width=0.99\textwidth]{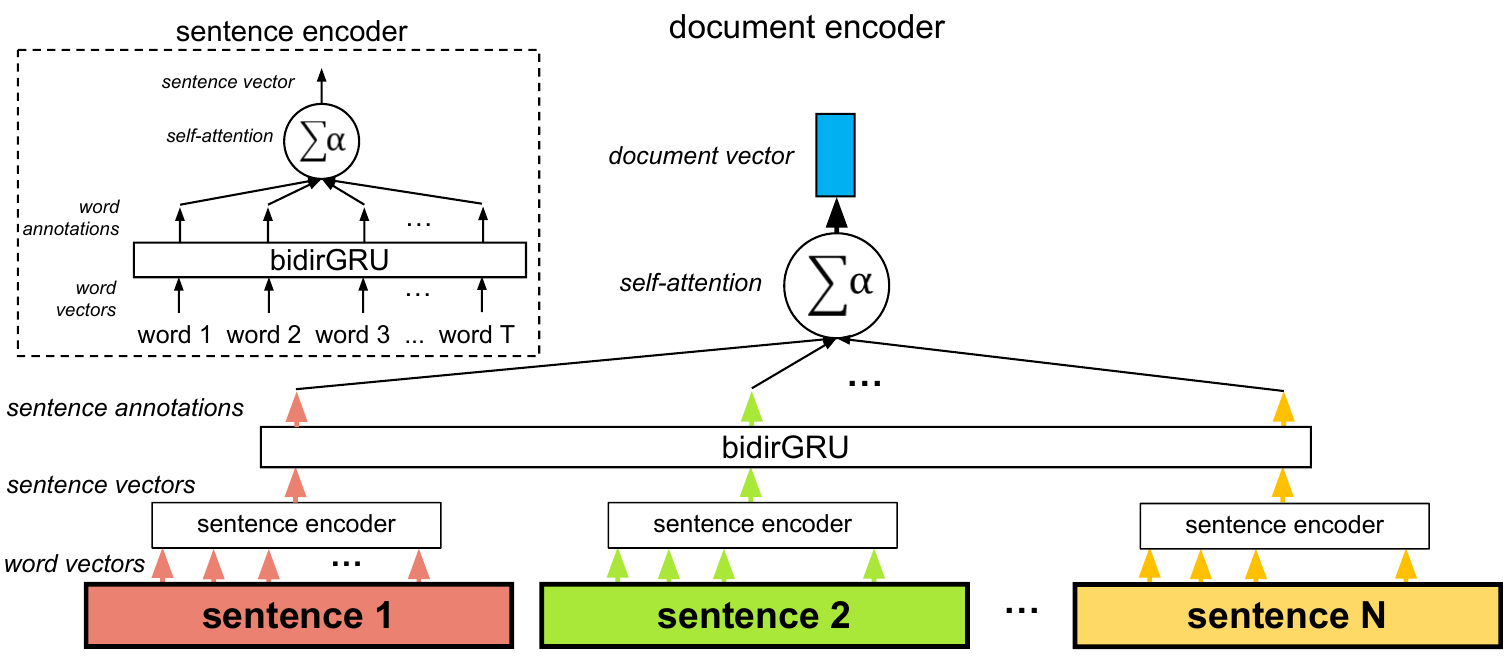}
\captionsetup{size=small}
\caption{Hierarchical Attention Network (HAN), reproduced with permission from \cite{tixier2018notes}.}\label{fig:han}
\end{figure}

A famous example is HAN \cite{yang2016hierarchical}, shown in Fig. \ref{fig:han}. In HAN, each sentence is first separately encoded. During this process, the model determines which words are important in each sentence by using a self-attention mechanism. Second, a representation for the full document is obtained from the sentence vectors. During this second step, the model is able to determine which sentences are the most important through the use of another self-attention mechanism. These two levels of attention enable the model to capture the fact that a given instance of a word may be very important when found in a given sentence, but that another instance of the same word may be less important when found in another sentence. The sentence and document encoder share the same architecture: a bidirectional Recurrent Neural Network (RNN) with Gated Recurrent Units (GRU). However, they do not share the same weights. In what follows, we briefly explain the different elements that compose the HAN architecture.

\noindent \textbf{Recurrent Neural Networks (RNNs)}.
CNNs are able to capture word order locally within each instantiation of the sliding window, but they ignore long-range dependencies \cite{goldberg2016primer}. On the other hand, RNNs were specifically developed to be used with \textit{sequences} \cite{elman1990finding}. From a high level, a RNN is fed an ordered list of input vectors $\big\{x_{1},...,x_{T}\big\}$ and returns an ordered list of \textit{hidden states} or \textit{annotations} $\big\{h_{1},...,h_{T}\big\}$. At any step $t$ in the sequence, the hidden state $h_{t}$ is defined in terms of the previous hidden state $h_{t-1}$ and the current input vector $x_{t}$ 

\begin{equation}\label{eq:rnn_hidden}
h_{t} = \mathrm{unit}(x_{t},h_{t-1})
\end{equation}

\noindent A \textbf{bidirectional RNN} is made of two unidirectional RNNs, that do not share the same parameters. The first \textit{forward} RNN processes the source sentence from left to right, while the second \textit{backward} RNN processes it from right to left. The two annotations are concatenated at each time step $t$ to obtain the bidirectional annotation:

\vspace{-0.5cm}

\begin{equation}\label{eq:bi_gru}
h_{t} = \big[ \vec{h_t}; \cev{h_t} \big]
\end{equation}

\noindent The bidirectional RNN takes into account the entire context when encoding the source words, not just the preceding words. As a result, $h_t$ is biased towards a small window centered on word $x_t$, while with a unidirectional RNN, $h_t$ is biased towards $x_t$ and the words immediately preceding it.

\noindent \textbf{GRU}. The basic RNN unit has been significantly improved throughout the years. Notably, the Long Short Term Memory unit (LSTM) \cite{hochreiter1997long} and the Gated Recurrent Unit (GRU) \cite{cho2014learning} made major strides, allowing RNN architectures to keep track of information over longer periods of time. The GRU is a simplified LSTM unit. While both the LSTM and GRU units are clearly superior to the basic RNN unit, there is no evidence about which one is best \cite{greff2017lstm,chung2014empirical}. We used the GRU in our experiments since it is more simple, more efficient, and has less parameters (so reduces the risk of overfitting).

\noindent \textbf{Self-attention}. The attention mechanism was developed in the context of encoder-decoder architectures for Machine Translation \cite{sutskever2014sequence,bahdanau2014neural}. Today, attention is ubiquitous in deep learning. When used in an encoder-only context, such mechanisms are qualified as \textit{self} or \textit{inner} attention \cite{lin2017structured,yang2016hierarchical}. Self-attention removes the need for the RNN encoder to try to fit information about the entire input sequence into its last hidden state at the risk of discarding a lot of information. Rather, a new vector (the attentional vector) summarizing the entire input sequence is computed by taking into account the annotations at \textit{all} time steps. This allows the encoder to distribute important information about the input into all its annotations. Therefore, much more information is kept. 

In our experiments, we used the self-attention mechanism as described in \cite{yang2016hierarchical}. As shown in Eq. \ref{eq:self-att}, annotation $h_t$ is first passed to a dense layer. An alignment coefficient $\alpha_t$ is then derived by comparing the output $u_t$ of the dense layer with a trainable context vector $u$ (initialized randomly) and normalizing with a softmax. Comparison is performed via a dot product. The attentional vector $\mathrm{att}$ is finally obtained as a weighted sum of all annotations.

\vspace{-0.4cm}

\begin{equation}\label{eq:self-att}
  \begin{split}
    u_t &= \mathrm{tanh}(Wh_t)\\
    \alpha_t &= \frac{\exp(u_{t}^\top u)}{\sum_{t'=1}^T \exp(u_{t'}^\top u)}\\
    \mathrm{att} &= \sum_{t=1}^T \alpha_t h_t
  \end{split}
\end{equation}

\noindent The context vector $u$ can be interpreted as a representation of the optimal word, on average. When faced with a new example, the model uses $u$ to decide which words it should pay attention to. During training, the model adjusts its internal representation of what the optimal word is (i.e., $u$ is updated) so as to maximize downstream performance.

\noindent \textbf{Truncation and padding}. By inspecting the distributions of the number of words per sentence and number of sentences per case, we set the maximum length of a sentence to be 50 words, and the maximum number of sentences in a document to be 14. Longer (respectively, shorter) sentences/documents were truncated (respectively, zero-padded).

\subsection{Softmax layer}
For both deep learning architectures, since our objective was \textit{classification}, we passed the document vector to a final dense layer with a softmax activation function. The softmax transforms a vector $x \in \mathbb{R}^{K}$ into a vector of positive floats that sum to one, i.e., into a \textit{probability distribution} over the $K$ classes to be predicted:

\begin{equation}
\mathrm{softmax}(x_{i}) = \frac{e^{x_{i}}}{\sum_{j=1}^{K}e^{x_{j}}}
\end{equation}

\noindent Note that in our experiments, the number of categories $K$ was different for each outcome, so the final softmax layers of the CNN and HAN architectures were not exactly the same for each outcome. The training and test sets were different for each outcome as well. In the end, we therefore constructed and trained 8 different models (4 outcomes $\times$ 2 architectures).

\subsection{TF-IDF + SVM}\label{sub:svm}
\noindent \textbf{TF-IDF}. The BoW representation, or vector space, has already been presented in subsection \ref{sub:background}. Here, we briefly present the standard weighting scheme Term Frequency-Inverse Document Frequency (TF-IDF). TF-IDF turns the non-zero coefficients of the BoW vectors into real weights with the assumption that frequent words in a document are representative of that document as long as they are not also very frequent at the corpus level. More precisely, the weights are computed as:

\vspace{-0.5cm}

\begin{equation}
\mathrm{weight}(t,d) = tf(t,d) \times idf(t,D)
\end{equation}

\noindent where $tf(t,d)$ is the number of times term $t$ appears in document $d$, and \linebreak $idf(t,D) = \log \big( \nicefrac{m+1}{df(t)} \big)$, with $df(t)$ the number of documents in the collection $D$ that contain $t$ and $m$ the size of $D$. We set the maximum document size to 200, like for the CNN model.

\begin{figure}[h]
\centering
\includegraphics[width=0.45\textwidth]{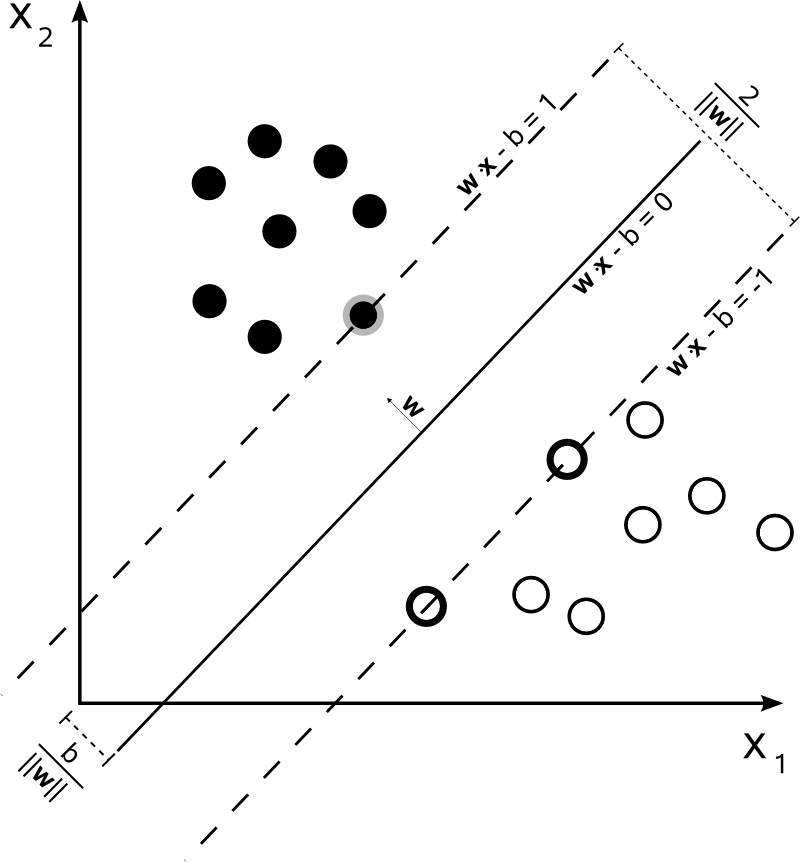}
\captionsetup{size=footnotesize}
\caption{Linear SVM decision boundary in the two-dimensional case. \label{fig:svm}}
\end{figure}

\noindent \textbf{SVM}.
In our experiments, we used a linear Support Vector Machine \cite{boser1992training}. The linear SVM is a geometrical method that seeks to classify points into two different categories by finding the best separating hyperplane. As illustrated by Fig. \ref{fig:svm} in the two-dimensional case, the best separating hyperplane is the line that separates the two groups of points with the greatest possible margin on each side. Training the SVM, that is, finding the best hyperplane, comes down to optimizing the $\vec{w}$ and b parameters (see Fig. \ref{fig:svm}). In test mode, a new observation is classified based on the side of the hyperplane on which it falls, which corresponds to the sign of the dot product of the observation with the vector orthogonal to the hyperplane ($\vec{w}$).

\noindent Because in practice, points may not all be  separable (e.g., due to outliers), when searching for the best separating hyperplane, the SVM is allowed to misclassify certain points. The tolerance level is controlled by a parameter traditionally referred to as C in the literature. The smaller C, the more tolerant the model is towards misclassification. C plays a crucial \textit{regularization} role, i.e., it has a strong impact on the generalization ability of the SVM. Indeed, for large values of C (low misclassification tolerance), a smaller-margin hyperplane will be favored over a larger-margin hyperplane if the former classifies more points correctly, at the risk of overfitting the training data. On the other hand, small values of C will favor larger-margin separating hyperplanes, even if they misclassify more points. Such solutions tend to generalize better.

When the target variable features more than two categories, a \textit{one-versus-rest} approach is used to redefine the problem as a set of binary classification tasks. More precisely, as many SVMs as there are categories are trained, where the goal of each SVM is to predict whether an observation belongs to its associated category or not.

\section{Dataset}

\subsection{Text and outcomes}
We had access to a large dataset of more than 90,000 incident reports provided by a global industrial partner in the oil and gas sector. The reports corresponded to work done on five continents from the early 2000s to 2018.

For each incident report, at least one of the following four text fields were available: (1) \texttt{title}: provides a quick summary of the report, (2) \texttt{description}: full description of the incident, (3) \texttt{details}: gives some details about the physical environment in which the incident occurred (e.g., ``engine room'', ``7$^{th}$ floor'', ``main deck''), (4) \texttt{root\_cause}: tries to isolate the root cause(s) of the incident (e.g., ``untidy housekeeping'').

About three fourths of the reports were originally in English. The rest of them were in other languages and had been automatically translated to English using commercial software.
Also, due to the geographical dispersion and large time span of the database, reports had been entered in the database by hundreds of different people, so that they were of very different styles.

The following outcomes were available for some of the injury reports: (1) \texttt{severity}: severity of the injury, (2) \texttt{injury\_type}: nature of the injury, (3) \texttt{bodypart}: body part(s) impacted during the incident, and (4) \texttt{incident\_type}: nature of the incident. Each outcome was associated with a certain number of unique values, that we will denote in the remainder of this article interchangeably as categories, classes, or levels. Category counts for each outcome are shown in Table \ref{table:out_counts}.

\begin{table}[h]
\centering

\scalebox{0.7}{

\begin{tabular}[t]{llr}
\multicolumn{2}{c}{incident type} \\
\hline
eq./tools    & 29033 \\
slips/trips/falls & 20474 \\
rules         & 14296 \\
access        & 10368 \\
PPE           &  8998 \\
dropped           &  8469 \\
\end{tabular}

\begin{tabular}[t]{llr}
\multicolumn{2}{c}{injury type} \\ 
\hline
contusion     &   4048 \\
cut/puncture &   3737 \\
FOB         &   2073 \\
pain          &   1587 \\
\end{tabular}

\begin{tabular}[t]{llr}
\multicolumn{2}{c}{bodypart} \\
\hline
finger &   4421 \\
hand &   3161 \\
eye    &   2392 \\
lower extr.  &   2321 \\
head     &   2184 \\
upper extr.  &   1757 \\
\end{tabular}

\begin{tabular}[t]{llr}
\multicolumn{2}{c}{severity} \\
\hline
1st aid &  16510 \\
med./restr.  &   2875 \\
\end{tabular}

}

\captionsetup{size=footnotesize}
\caption{Category counts for each outcome. \label{table:out_counts}}
\end{table}

\subsection{Text preprocessing}\label{sub:dl-prepro}

\noindent Standard text preprocessing steps were applied to the text fields of each report. This included tokenization into sentences and words, and conversion to lowercase. We also removed non-ASCII characters and HTML formatting, and performed some other fine-cleaning steps. Finally, we used a word segmenter to correct some of the many segmentation issues present in the text fields (``fellfrom'' $\rightarrow$ ``fell from'').

\subsection{Train/test splits and class imbalance}
We created separate train/test splits for each outcome: 90\% of the reports were used for training, and 10\% were left-out for testing. Categories were not mutually exclusive for \texttt{injury\_type} and \texttt{bodypart}. For these outcomes, 301 and 519 reports (respectively) were associated with more than one level. In order to make use of as much data as possible, we considered these reports to belong to \textit{all} the categories they were associated with, rather than arbitrarily selecting a single one. However, to make sure that no report from the training sets leaked to the test sets, such extension was performed only after the initial train/test splits had been made. 

To address the problem of class imbalance, weights inversely proportional to the frequency of each category in the training set were computed. During training only, these weights were used to weigh the loss function and ensure that the model paid equal attention to all levels, and in particular, did not neglect the minority categories. Statistics about the final train/test splits are shown in Tables \ref{table:tt_splits_1} to \ref{table:tt_splits_4}.

\begin{table}[h]
\centering
\setlength{\tabcolsep}{2pt}
\scalebox{0.7}{
\begin{tabular}[]{rcccccc|cccccc}
 & \multicolumn{6}{c|}{train (82,474)} 
& \multicolumn{6}{c}{test (9,164)} \\
\hline
 & access & eq./tools & slips/trips/falls & dropped & PPE & rules & access & eq./tools & slips/trips/falls & dropped & PPE & rules \\ 
 counts & 9,300 & 26,167 & 18,432 & 7,619 & 8,078 & 12,878 & 2,866 & 850 & 1,068 & 2,042 & 920 & 1,418 \\ 
 weights & 2.8 & 1.0 & 1.4 & 3.4 & 3.2 & 2.0  &  - & -  & -  & -  & -  & -  
\end{tabular}
}
\captionsetup{size=footnotesize}
\caption[Train/test splits statistics for \texttt{incident\_type}]{Train/test splits statistics for \texttt{incident\_type}.\label{table:tt_splits_1}}
\end{table}

\begin{table}[h]
\centering
\setlength{\tabcolsep}{2pt}
\scalebox{0.7}{
\begin{tabular}[]{rcccc|cccc}
& \multicolumn{4}{c|}{train (10,299)} 
& \multicolumn{4}{c}{test (1,146)} \\  
\hline
& cut/puncture & FOB & pain & contusion & cut/puncture & FOB & pain & contusion \\ 
 counts & 3,355 & 1,884 & 1,443 & 3,617 & 382 & 189 & 431 & 144 \\ 
 weights & 1.1 & 1.9 & 2.5 & 1.0  & -  & -   & -  & -   
\end{tabular}
}
\captionsetup{size=footnotesize}
\caption[Train/test splits statistics for \texttt{injury\_type}]{Train/test splits statistics for \texttt{injury\_type}.\label{table:tt_splits_2}} 
\end{table}

\begin{table}[h]
\centering
\setlength{\tabcolsep}{2pt}
\scalebox{0.7}{
\begin{tabular}[]{rcccccc|cccccc}
 & \multicolumn{6}{c|}{train (14,616)} 
& \multicolumn{6}{c}{test (1,620)} \\
\hline
 & hand & lower extr. & upper extr. & head & finger & eye & hand & lower extr. & upper extr. & head & finger & eye \\ 
 counts & 2,821 & 2,102 & 1,583 & 1,958 & 4,031 & 2,121 & 340 & 174 & 219 & 226 & 390 & 271 \\ 
 weights & 1.4 & 1.9 & 2.5 & 2.1 & 1.0 & 1.9  &  - & -  & -  & -  & -  &  -
\end{tabular}
}
\captionsetup{size=footnotesize}
\caption[Train/test splits statistics for \texttt{bodypart}]{Train/test splits statistics for \texttt{bodypart}.\label{table:tt_splits_3}}
\end{table}

\begin{table}[h]
\centering
\setlength{\tabcolsep}{2pt}
\scalebox{0.7}{
\begin{tabular}[]{rcc|cc}
& \multicolumn{2}{c|}{train (17,447)} 
& \multicolumn{2}{c}{test (1,938)} \\  
\hline
& 1st aid & med./restr. & 1st aid & med./restr. \\ 
counts & 14,860 & 2,587 & 1,650 & 288 \\ 
weights & 1.0 & 5.7  & -  & -    
\end{tabular}
}
\captionsetup{size=footnotesize}
\caption{Train/test splits statistics for \texttt{severity}.\label{table:tt_splits_4}}
\end{table}

\subsubsection{Vocabulary construction}
Reports that did not belong to the test set of any outcome were used to build the vocabulary and pre-train word embeddings. There were more than 218K such reports.
First, the number of occurrences of each unique token in the corpus was computed. Then, the tokens appearing at least 5 times were retained, which gave us about 24K tokens. The selected tokens were then sorted by decreasing order of frequency and assigned integer values from 2 (most frequent) to about 24K (least frequent). This dictionary was our vocabulary. It was used to convert the text parts of the reports into arrays of integers suited to be passed to the models. The indexes 0 and 1 were reserved respectively for zero-padding and for out-of-vocabulary words (i.e., appearing less than 5 times, or appearing only in the test set).

\subsubsection{Pre-training word vectors}
Transfer learning is best harnessed when the domain of pre-training and that of the downstream task are not too different.
On the other hand, when the data originate from a highly specific domain, and when sufficient amounts of them are available, it is considered better practice that one pre-trains its own word vectors on their own dataset \cite{yang2016hierarchical}. 
Therefore, since incident reports are very specific, and since our dataset was big enough to allow effective pre-training, we used the gensim \cite{rehurek_lrec} implementation of \texttt{word2vec} (version \texttt{3.2.0.}) to learn 64-dimensional word vectors on our training sets, with the skip-gram variant. Table \ref{table:wv_san} shows the most similar words to some anchor words in the learned embedding space, for sanity checking. Clearly, similarities are sensical.

\begin{table}[h]
\centering
\scalebox{0.7}{

\begin{tabular}[t]{llr}
\multicolumn{2}{c}{\textbf{worker}} \\
\hline
employee &  83.1 \\
workman & 80.53 \\
victim & 80.04 \\
ip & 79.97 \\
fitter & 79.57 \\
fabricator & 78.73 \\

\end{tabular}

\begin{tabular}[t]{llr}
\multicolumn{2}{c}{\textbf{pipe}} \\ 
\hline
pipes & 81.25 \\
coupon & 77.64 \\
pup & 75.77 \\
pipeline & 75.49 \\
joint & 73.22 \\
stalk & 72.51 \\

\end{tabular}

\begin{tabular}[t]{llr}
\multicolumn{2}{c}{\textbf{valve}} \\
\hline
valves & 86.59 \\
bleeder & 79.22 \\
manifold & 76.88 \\
solenoid & 76.61 \\
actuator & 76.46 \\
inlet & 75.64 \\

\end{tabular}

\begin{tabular}[t]{llr}
\multicolumn{2}{c}{\textbf{hammer}} \\
\hline
sledge & 90.01 \\
sledgehammer & 84.29 \\
spanner & 83.29 \\
chisel & 82.33 \\
wrench & 82.11 \\
mallet & 80.68 \\
\end{tabular}
}
\captionsetup{size=footnotesize}
\caption[Word vectors sanity checks]{Word vectors sanity checks. Rows correspond to the 6 words most similar to the column header (in terms of cosine similarity).\label{table:wv_san}}
\end{table}

\section{Experimental setup}

\subsection{Deep learning models}
\noindent \textbf{Loss function}. The function that we minimized during training was the categorical cross-entropy, also known as the log loss:

\vspace{-0.3cm}

\begin{equation}
\mathrm{logloss} = -\frac{1}{N}\sum_{i=1}^{N}\sum_{j=1}^{K}y_{ij}\mathrm{log}p_{ij}
\end{equation}

\noindent Where $N$ is the number of reports in the training set, $K$ is the number of categories, $y_{ij}$ is the one-hot label vector of the $i^{th}$ example (size $K$, zero everywhere, except for the index corresponding to the category of the example), and $p_{ij}$ is the probability returned by the model that the $i^{th}$ example falls into the $j^{th}$ category. Since $y_{ij}$ is equal to zero everywhere except for $j$ corresponding to the index of the true label, only the probability assigned by the model to the true category of the $i^{th}$ example contributes to the overall score.

Let us consider a report with label vector $[0,1]$. Imagine that the probabilistic forecast of the model is $[0.2,0.8]$, which corresponds to a quite good prediction. The log loss will be equal to $-\mathrm{log}(0.8)=0.22$. If the prediction is slightly worse, but not completely off, say $[0.4,0.6]$, the log loss will be equal to $0.51$, and for $[0.9,0.1]$ (very bad prediction), the log loss will reach $2.3$. We can observe that the further away the model gets from the truth, the greater it gets penalized. For a perfect prediction (probability of 1 for the right class), the categorical cross-entropy is null and the model receives no penalty.

\noindent \textbf{Cyclical Learning Rate and Momentum}. To minimize the loss during training, we used the Stochastic Gradient Descent (SGD) optimizer with opposite triangle policy Cyclical Learning Rate (CLR) and Cyclical Momentum (CM) schedules, as described in \cite{smith2017cyclical,smith2018disciplined}. We opted for SGD as there is more and more evidence that adaptive optimizers like Adam, Adagrad, etc. converge faster but generalize poorly compared to SGD\footnote{\tiny{\url{https://shaoanlu.wordpress.com/2017/05/29/sgd-all-which-one-is-the-best-optimizer-dogs-vs-cats-toy-experiment/}}} \cite{wilson2017marginal}. It is true that SGD is much slower than adaptive approaches, but the use of cyclical learning rates can bring a significant speedup, and even sometimes allow to reach better performance.

Following the recommendations of \cite{smith2017cyclical,smith2018disciplined}, we estimated the bounds between which the learning rate should vary with a \textit{learning rate range test}. We conducted such a test for each outcome, with half cycle of 6 epochs. The maximum value of the learning rate was then selected as the point right before the plateau. The minimum value was set as the maximum value divided by 6. Note that for each test, the loss and the performance metrics were computed on a \textit{validation} set that we constructed by randomly sampling 11.1\% of observations at random from the training set, without replacement, so as to obtain a validation set of roughly same size as the test set. The final ranges used for training are shown in Table \ref{table:lr_range_test} for each architecture and each outcome.

\begin{table}[h]
\centering
\scalebox{0.6}{
\begin{tabular}[t]{rcccccccc}
& \multicolumn{2}{c}{incident type} & \multicolumn{2}{c}{injury type} & \multicolumn{2}{c}{bodypart} & \multicolumn{2}{c}{severity} \\
\hline
& min & max & min & max & min & max & min & max \\
range test & $10^{-6}$ & $4.5\times7\times 10^{-2}$ & $10^{-6}$ & $7\times10^{-2}$ & $10^{-6}$ & $7\times10^{-2}$ & $10^{-6}$ & $7.\times10^{-2}$ \\
train HAN &$ 1.04\times10^{-2} $&$ 6.29\times10^{-2} $&$ 3.1\times10^{-3} $&$ 1.86\times10^{-2} $&$ 4.85\times10^{-3} $&$ 2.91\times10^{-2} $&$ 7.72\times10^{-4} $&$ 4.63\times10^{-3} $ \\
train CNN &$ 4.37\times10^{-3} $&$ 2.62\times10^{-2} $&$ 1.16\times10^{-3} $&$ 6.99\times10^{-3} $&$ 2.71\times10^{-3} $&$ 1.63\times10^{-2} $&$ 1.94\times10^{-4} $&$ 1.16\times10^{-3}$
\end{tabular}
}
\captionsetup{size=footnotesize}
\caption{Learning rate ranges explored during the LR range tests and used for training. \label{table:lr_range_test}}
\end{table}

\noindent No range test is necessary for Cyclical Momentum \cite{smith2018disciplined}. We simply used the recommended values of 0.95 and 0.85 in our experiments.

\noindent \textbf{Regularization}. For regularization, we used dropout \cite{srivastava2014dropout} at all layers and an early stopping strategy based on the test loss.

\subsection{TF-IDF + SVM}
We considered up to trigrams as dimensions of the BoW space (i.e., $n$-grams with $n \in [1,3]$). The maximum number of dimensions was set to be 6 times the size of the deep learning vocabulary. The terms that were occurring in more than 90\% of the reports were ignored.

The C parameter of the linear SVM was tuned for each outcome with a grid search on a validation set of same size as that used for the learning rate range tests.
We tried 24 equally spaced values, from $x=-5$ to $x=6.5$, where C=$10^x$. After the grid search, a final model was trained on the training and validation sets with the optimal value of C, and tested on the test set.

\subsection{Performance metrics}
The same performance metrics were used for all models. Due to the large class imbalance for all outcomes, measuring classification performance with accuracy was inadequate. Rather, we recorded a \textit{confusion matrix} at the end of each batch on the validation set during the learning rate range tests and on the test set during training. The confusion matrix $C$ is a square matrix of dimension $K \times K$ where $K$ is the number of categories, and the $(i,j)^{th}$ element $C_{i,j}$ of $C$ indicates how many of the observations known to be in category $i$ were predicted to be in category $j$. From the confusion matrix, we computed precision, recall and F1-score for each class. Precision, respectively recall, for category $i$, was computed by dividing $C_{i,i}$ (the number of correct predictions for category $i$) by the sum over the $i^{th}$ column of $C$ (the number of predictions made for category $i$), respectively by the sum over the $i^{th}$ row of $C$ (the number of observations in category $i$).

\vspace{-0.5cm}

\begin{align}
\text{precision} & = \frac{C_{i,i}}{\sum_{j=1}^{K}C_{j,i}} \\
\text{recall} & = \frac{C_{i,i}}{\sum_{j=1}^{K}C_{i,j}}
\end{align}

\noindent Finally, the F1-score was computed classically as the harmonic mean of precision and recall.

\vspace{-0.5cm}

\begin{align}
\text{F1} & = 2 \times \frac{\text{precision} \times \text{recall}}{\text{precision} + \text{recall}} 
\end{align}

\subsection{Configuration}
Experiments were run in Python 3.6 with \texttt{tensorflow-gpu} \cite{abadi2016tensorflow} version 1.5.0 and \texttt{Keras} \cite{chollet2015keras} front-end version 2.2.0, using a high-end GPU. For both TF-IDF vectorization and SVM classification, we used the \texttt{scikitlearn} library \cite{scikitlearn}. For the TF-IDF + SVM approach, preprocessing and the train/test splits were the same as for the deep learning architectures.

\section{Classification performance}
Results can be seen in Tables \ref{table:res_case_cat} to \ref{table:res_acc_cat}. The best F1 score for each column is indicated in \textbf{bold}. Performance is always much better than random, and very high scores are reached for some categories. For instance, the FOB (foreign object) class of the \texttt{injury\_type} outcome is predicted with 95.49 F1 score by CNN, while the eye category of the {\small\texttt{bodypart}} outcome is predicted almost perfectly by HAN (97.04 F1). Even for the categories that are more difficult to predict for all models, such as Rules for \texttt{incident\_type} or Pain for \texttt{injury\_type}, best scores reach a decent level (61.75 and 66.17 resp., both attained by TF-IDF+SVM).\\

\noindent \textbf{HAN vs CNN}. Overall, while being slightly more parsimonious (see Table \ref{table:time_epoch}), HAN outperforms CNN almost everywhere. The only exceptions are FOB for \texttt{injury\_type} and upper extr. for \texttt{bodypart}. This performance superiority is probably due to the fact that HAN features an attention mechanism. It could also mean that capturing word order outside of small local windows is beneficial. However, training takes three times longer, on average, for HAN than for CNN. This is due to the fact that even high performance CUDA optimized implementations of recurrent operations take much less advantage of the GPU than convolutional operations.

\begin{table}[H]
\centering
\scalebox{0.7}{
\begin{tabular}[t]{rcccccccc}
 task & \multicolumn{2}{c}{incident type} & \multicolumn{2}{c}{injury type} & \multicolumn{2}{c}{bodypart} & \multicolumn{2}{c}{severity} \\
\hline
model & HAN & CNN & HAN & CNN & HAN & CNN & HAN & CNN \\
time/ep. & 216.99 & 71.47 & 32.31 & 11.54 & 46.33 & 16.46 & 54.39 & 19.76 \\
\#params & 1,631,966 & 1,709,666 & 1,551,844 & 1,589,664 & 1,551,926 & 1,590,266 & 1,551,762 & 1,589,062
\end{tabular}
}
\captionsetup{size=footnotesize}
\caption[Runtime per epoch and number of parameters]{Average runtime per epoch (in secs) on a high-end GPU and number of parameters for the deep learning models. \label{table:time_epoch}}
\end{table}

\noindent \textbf{TF-IDF+SVM vs deep learning}. TF-IDF+SVM lives up to its reputation, and reaches very high performance everywhere. Interestingly, it even outperforms deep learning, except for \texttt{bodypart} (see  Table \ref{table:res_inj_bod}). We hypothesize that deep learning is not able to stand out on our relatively small datasets (for deep learning's standards). On very large datasets, deep learning is expected to be significantly better than TF-IDF+SVM. See, for instance, Table 2 in \cite{yang2016hierarchical}.

While TF-IDF+SVM is very strong on our datasets, we definitely cannot conclude that deep learning should be abandoned in future research.
Indeed, detecting the presence or absence of some specific patterns (keywords and keyphrases) was apparently sufficient to perform well at our classification tasks.
In that case, the superior expressiveness of deep learning over more simple models cannot show.
However, there are applications that require a deeper understanding of the input documents and the ability to generate text, such as translation and summarization, in which TF-IDF cannot compete with deep learning or is just not even applicable.

\renewcommand{\arraystretch}{0.85} 
\begin{table}[H]
\centering
\scalebox{0.75}{
\begin{tabular}{rrcccccccc}
&      & access & dropped  & eq./tools & PPE   & rules & slips/trips/falls & mean  \\
\hline
 HAN & prec & 53.22   & 67.21 & 76.3        & 71.73 & 59.62 & 77.68          & 67.62 \\
 HAN & rec  & 70.51   & 77.88 & 63.47       & 85.76 & 57.48 & 72.77          & 71.31 \\
 HAN & F1   & 60.65   & 72.15 & 69.3        & 78.12 & 58.53 & 75.14          & 68.98 \\
\hline
 CNN & prec & 59.56   & 81.5  & 56.16       & 77.28 & 53.44 & 80.99          & 68.15 \\
 CNN & rec  & 50.47   & 60.12 & 82.03       & 72.83 & 43.86 & 56.12          & 60.9  \\
 CNN & F1   & 54.64   & 69.19 & 66.68       & 74.99 & 48.18 & 66.3           & 63.33 \\
 \hline
   TF-IDF+SVM & prec &  62.81   & 72.18 & 74.31       & 77.9  & 65.79 & 75.52          & 71.42 \\
 TF-IDF+SVM & rec  & 63.11   & 76.0  & 75.68       & 81.63 & 58.18 & 76.15          & 71.79 \\
 TF-IDF+SVM & F1   & \textbf{62.96}   & \textbf{74.04} & \textbf{74.99} & \textbf{79.72} & \textbf{61.75} & \textbf{75.83} & \textbf{71.55} \\
 \hline
random & prec & 16.95 & 16.82 & 16.61 & 17.07 & 16.64 & 15.67 & 16.63 \\ 
random & rec & 11.98 & 9.29 & 30.89 & 10.44 & 15.23 & 21.08 & 16.48 \\ 
random & F1 & 14.04 & 11.97 & 21.60 & 12.95 & 15.90 & 17.98 & 15.74 \\ 
\end{tabular}
}
\captionsetup{size=footnotesize}
\caption[DL: \texttt{incident\_type} test set performance]{Test set performance for \texttt{incident\_type}. Best epochs are 35/60 (HAN/CNN). \label{table:res_case_cat}}
\end{table}

\vspace{-0.7cm}

\begin{table}[H]
\centering
\scalebox{0.75}{
\begin{tabular}{rrccccc}
     &      & contusion & cut/puncture & FOB   & pain    & mean  \\
     \hline
 HAN & prec & 85.49   & 87.74       & 90.69 & 51.96 & 78.97 \\
 HAN & rec  & 75.17   & 82.46       & 97.88 & 73.61 & 82.28 \\
 HAN & F1   & 80.00    & \textbf{85.02}       & 94.15 & 60.92 & 80.02 \\ 
\hline
 CNN & prec & 69.84   & 91.81       & 95.74 & 68.52 & 81.48 \\
 CNN & rec  & 90.26   & 70.42       & 95.24 & 51.39 & 76.83 \\
 CNN & F1   & 78.75   & 79.70        & \textbf{95.49} & 58.73 & 78.17 \\
 \hline
   TF-IDF+SVM & prec &  82.6    & 83.54       & 93.85 & 71.2  & 82.8  \\
 TF-IDF+SVM & rec & 82.6    & 86.39       & 96.83 & 61.81 & 81.91 \\
 TF-IDF+SVM & F1   &  \textbf{82.6}    & 84.94       & 95.32 & \textbf{66.17} & \textbf{82.26} \\
 \hline
random & prec & 20.89 & 26.44 & 21.16 & 22.22 & 22.68 \\ 
random & rec & 34.10 & 35.94 & 13.51 & 10.56 & 23.53 \\ 
random & F1 & 25.91 & 30.47 & 16.49 & 14.32 & 21.80 \\
\end{tabular}
}
\captionsetup{size=footnotesize}
\caption[DL: \texttt{injury\_type} test set performance]{Test set performance for \texttt{injury\_type}. Best epochs are 78/77 (HAN/CNN). \label{table:res_inj_type}}
\end{table}

\vspace{-0.7cm}

\begin{table}[H]
\centering
\scalebox{0.75}{
\begin{tabular}{rrcccccccc}
     &      & eye   & finger & head    & hand & lower extr. & upper extr. & mean  \\
     \hline
 HAN & prec & 97.4  & 88.0   & 89.62 & 87.63  & 85.07 & 73.06 & 86.8  \\
 HAN & rec  & 96.68 & 90.26  & 84.07 & 77.06  & 85.84 & 91.95 & 87.64 \\
 HAN & F1   & \textbf{97.04} & \textbf{89.11}  & \textbf{86.76} & \textbf{82.00}   & \textbf{85.45} & 81.42 & \textbf{86.97} \\
\hline
 CNN & prec & 97.73 & 86.86  & 88.41 & 76.02  & 72.54 & 85.55 & 84.52 \\
 CNN & rec  & 95.2  & 77.95  & 80.97 & 76.47  & 94.06 & 85.06 & 84.95 \\
 CNN & F1   & 96.45 & 82.16  & 84.53 & 76.25  & 81.91 & \textbf{85.30}  & 84.43 \\
 \hline
 TF-IDF+SVM & prec & 97.73 & 82.68  & 86.61 & 84.0   & 83.04 & 85.8  & 86.64 \\
 TF-IDF+SVM & rec  &  95.2  & 91.79  & 85.84 & 74.12  & 87.21 & 83.33 & 86.25 \\
 TF-IDF+SVM & F1   &  96.45 & 87.0   & 86.22 & 78.75  & 85.07 & 84.55 & 86.34 \\
  \hline
random & prec & 14.02 & 16.67 & 16.43 & 17.35 & 17.59 & 13.45 & 15.92 \\ 
random & rec & 14.73 & 24.44 & 12.59 & 19.93 & 15.08 & 9.16 & 15.99 \\ 
random & F1 & 14.37 & 19.82 & 14.26 & 18.55 & 16.24 & 10.90 & 15.69 \\  
\end{tabular}
}
\captionsetup{size=footnotesize}
\caption[DL: \texttt{bodypart} test set performance]{Test set performance for \texttt{bodypart}. Best epochs are 97/66 (HAN/CNN). \label{table:res_inj_bod}}
\end{table}

\begin{table}[H]
\centering
\scalebox{0.75}{
\begin{tabular}{rrccc}
     &      & 1st aid & med./restr. & mean \\
     \hline
 HAN & prec & 96.06    & 58.51   & 77.28 \\
 HAN & rec  & 90.24    & 78.82   & 84.53 \\
 HAN & f1   & 93.06    & 67.16  & 80.11 \\
  \hline
 TF-IDF+SVM & prec &    94.6     & 72.79   & 83.7  \\
 TF-IDF+SVM & rec  & 95.52    & 68.75   & 82.14 \\
 TF-IDF+SVM & F1   & \textbf{95.06}    & \textbf{70.71}  & \textbf{82.88} \\
  \hline
random & prec & 53.03 & 48.26 & 50.65 \\ 
random & rec & 85.45 & 15.21 & 50.33 \\ 
random & F1 & 65.44 & 23.13 & 44.29 \\ 
\end{tabular}
}
\captionsetup{size=footnotesize}
\caption[DL: \texttt{severity} test set performance]{Test set performance for \texttt{severity}. Best epoch is 103 for HAN. CNN not shown due to divergence (failed training). \label{table:res_acc_cat}}
\end{table}

\renewcommand{\arraystretch}{1} 

\begin{figure}[h]
\centering
\includegraphics[width=0.9\textwidth,trim={0 0 0 2.75cm},clip]{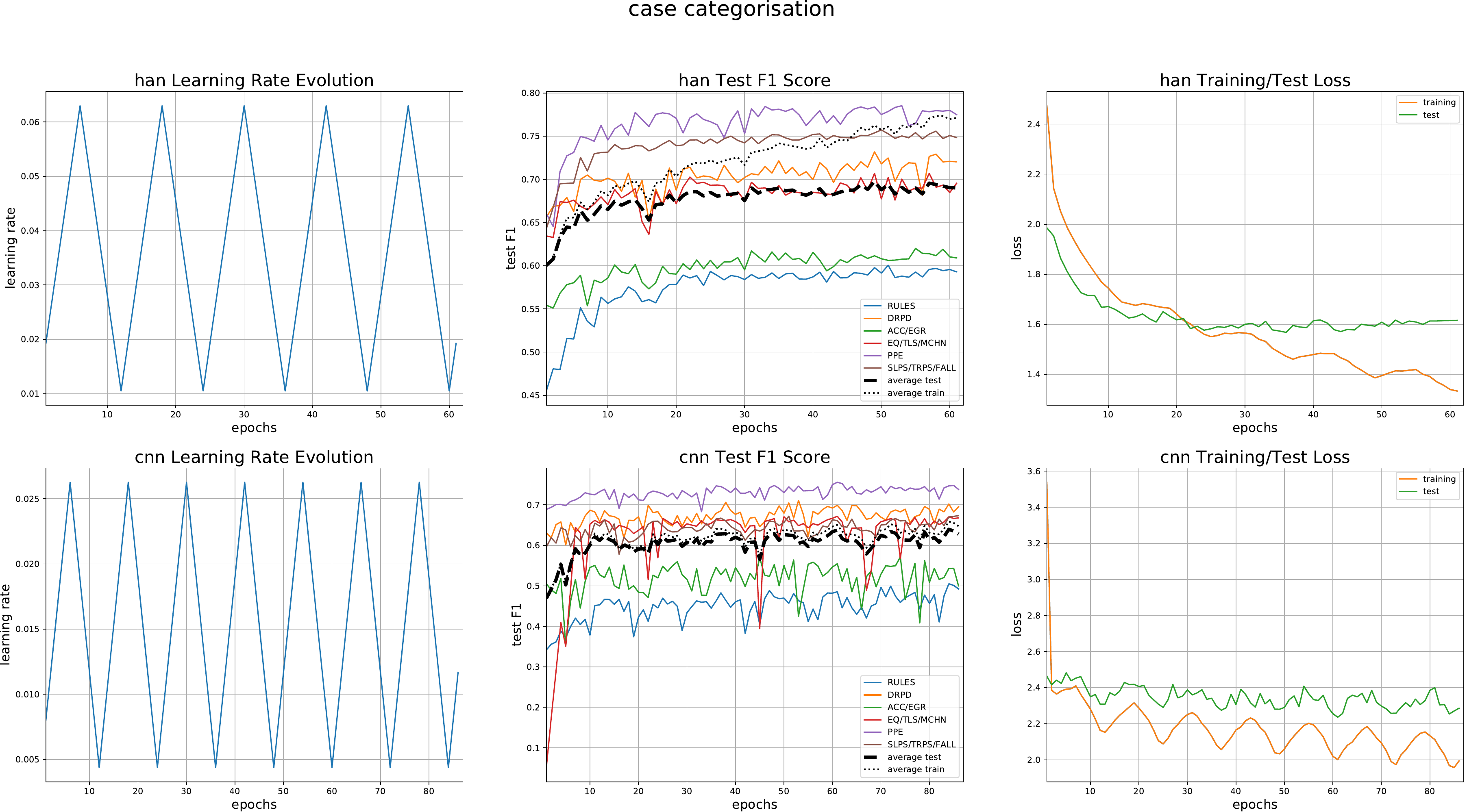}
\captionsetup{size=footnotesize}
\caption{Training history for \texttt{incident\_type}. \label{fig:res_case_cat}}
\end{figure}

\noindent \textbf{Sanity checks for deep learning}. We followed best practice\footnote{e.g., Stanford's CS231n course page, section ``Babysitting the learning process'': \url{http://cs231n.github.io/neural-networks-3/}} and inspected our loss and F1-score plots for any indicator of pathological training.
Except for the CNN model on the \texttt{severity} dataset which diverged after a few epochs, training was successful for all models and all outcomes.
In Fig. \ref{fig:res_case_cat} for instance, we can observe on the loss plot that we start overfitting after the 22th epoch, but the training and test loss curves only start diverging after the 30th epoch, and even then, the margin between the training and test loss curves is not too rapidly increasing. All of this indicates that the amount of overfitting is acceptable.
This suggests that the complexity of the models (i.e., their number of parameters) is appropriate and that our regularization strategy (dropout and varying learning rates) is effective, which enables the model to both fit the training data well and generalize well. Also, the loss curves seem neither too linear nor too exponential, which indicates that the learning rate is adequate.

\begin{figure}[h]
\centering
\includegraphics[width=0.457\textwidth,trim={0 0 0 0.45cm},clip]{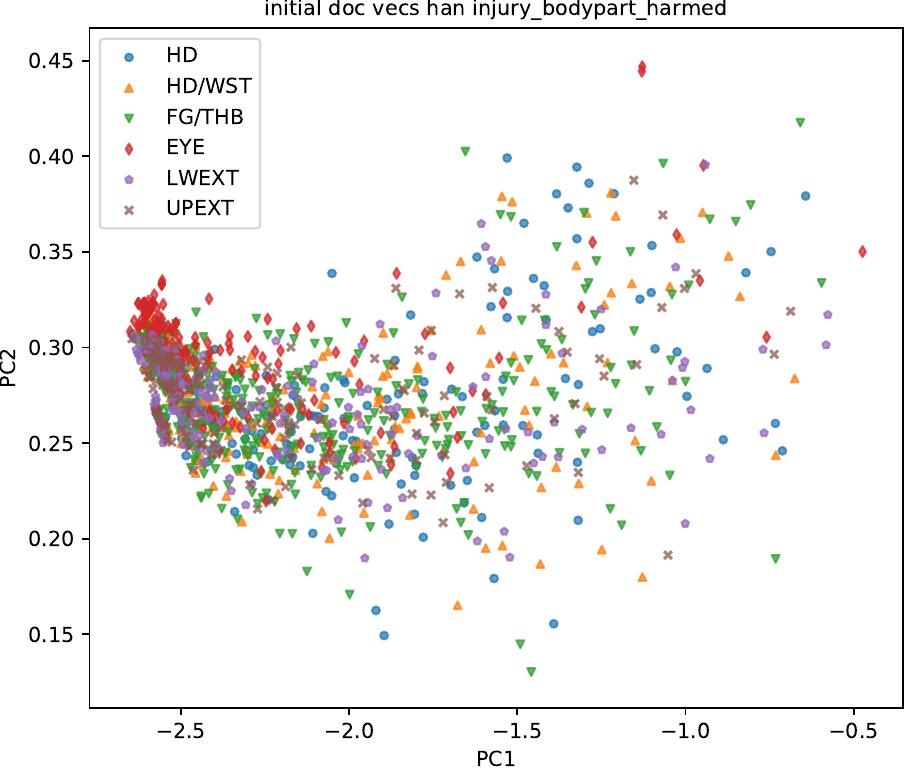}
\includegraphics[width=0.45\textwidth,trim={0 0 0 0.45cm},clip]{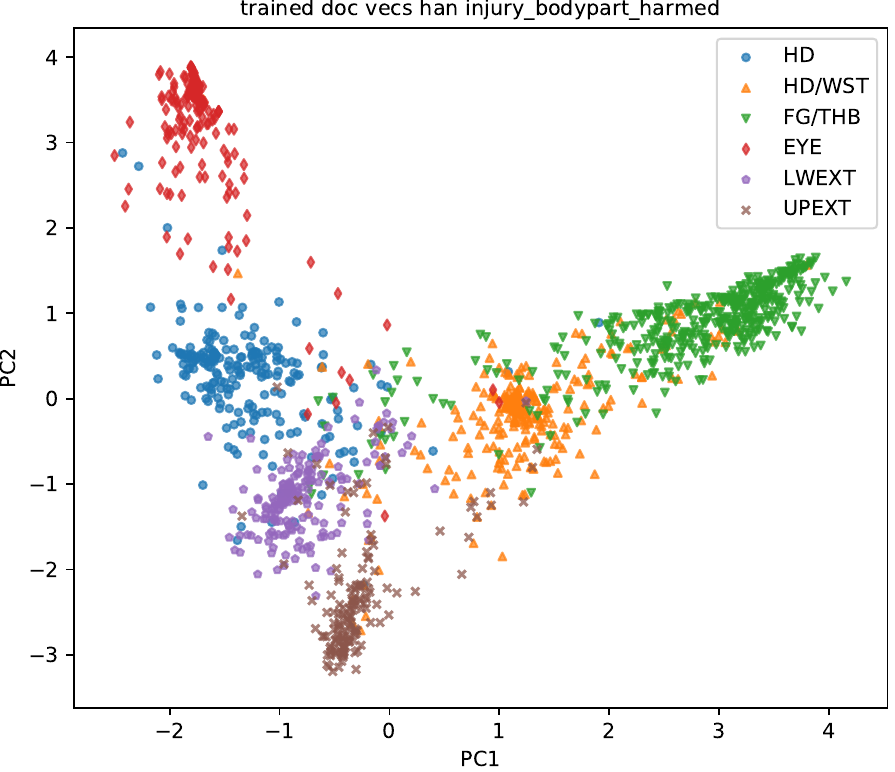}
\captionsetup{size=footnotesize}
\caption{HAN \texttt{bodypart} report embeddings \textit{before} training (left) and \textit{after} training (right). \label{fig:doc_emb_init}}
\end{figure}

Another easy way to verify that the deep learning models are learning effectively is to check whether their internal representations make sense after training. Fig. \ref{fig:doc_emb_init} provides a 2D PCA-projection of some randomly selected report embeddings.
We can observe that immediately after initialization, all reports are scattered at random in the internal space of the HAN \texttt{bodypart} model. This makes sense, since most of the weights of the model are initialized at random\footnote{The only parameters that are not initialized at random are the word embeddings.}. After training, however, we can see that the model has updated its weights in a way that the reports are now clearly organized in its internal space (see right of Fig. \ref{fig:doc_emb_init}). Indeed, reports are almost perfectly grouped based on the category they belong to, to the point that classes are almost linearly separable. 

In addition to sanity checking, such high-quality report embeddings can be used to perform semantic report retrieval, by passing a query to the model in test mode, that is, projecting the query in the internal space of the model, and getting  the nearest-neighbors. Such a search engine could be used to answer questions such as ``What incident reports in the database are the most similar to a given incident report?'', which could be very useful in practice.

\vspace{-0.5cm}

\section{Automatically learning incident precursors from text} \label{subsec:attimp}
In this section, we propose, for each model, a method to learn valid injury precursors from the raw dataset by using the trained models.
These precursors are identified among the parts of the raw input reports that we find the models rely on the most, on average, in predictive outcome categories.

Our methods can be used by construction management professionals to identify, in a fully automated and data-driven way, the most impactful hazards and behaviors, in order to better inform and focus safety efforts. The methods can also be used to better understand and interpret the model's predictions, and inspect the sections of each report that are most important in predicting each outcome. Transparency and trust in the methodology must be commensurate with predictive performance, if the models are to be adopted and used.

\subsection{CNN: Predictive regions and word saliency}

\noindent \textbf{Predictive regions}. This approach follows \cite{johnson2014effective} (section 3.6, Tables 5 \& 6). Recall that before we lose positional information through pooling, each of the $n_f$ filters of size $h$ is associated with a vector of size $s-h+1$ (a feature map) whose entries represent the output of the convolution of the filter with the corresponding region of the input. Therefore, each region is embedded into an $n_f$-dimensional space. Thus, after training, we can identify the regions of a given injury report that are the most predictive of its outcome level by inspecting the intermediate output of the model corresponding to the receptive field embeddings (right before the pooling layer), and by selecting the regions associated with the highest norm embeddings.

To identify the most predictive regions on average (i.e., at the dataset level), we simply fed all reports in the training set to the model in test mode and recorded the most predictive regions for each report. We applied this procedure to all outcomes. Table \ref{table:pred_regs} shows some of the results we obtained using this approach.

\begin{table}[h]
\centering
\begin{minipage}{0.32\textwidth}
\centering
\scalebox{0.75}{
\begin{tabular}{lr}
\multicolumn{2}{c}{Access (\texttt{incident\_type})} \\
\hline
 access to         & 110964 \\
 \textbf{trip hazard}       &  78240 \\
 the door          &  50170 \\
 \textbf{blocked by}        &  38836 \\
 access / egress   &  33280 \\
 door to           &  33050 \\
 \textbf{obstructed by}     &  28291 \\
 the access        &  22737 \\
 risk              &  19075 \\
 access and egress &  17171 \\
\hline
\end{tabular}
}
\end{minipage}
\begin{minipage}{0.32\textwidth}
\centering
\scalebox{0.75}{
\begin{tabular}{lr}
\multicolumn{2}{c}{Rules (\texttt{incident\_type})} \\
\hline
 safety glasses       & 38190 \\
 \textbf{not wearing}          & 36603 \\
 risk                 & 35970 \\
 \textbf{speeding in}          & 28067 \\
 \textbf{speeding in car park} & 22815 \\
 \textbf{smoking in}           & 21660 \\
 low risk             & 21658 \\
 dropped object       & 17788 \\
 \textbf{without wearing}      & 16024 \\
\hline
\end{tabular}
}
\end{minipage}
\begin{minipage}{0.32\textwidth}
\centering
\scalebox{0.75}{
\begin{tabular}{lr}
\multicolumn{2}{c}{Eye ( \texttt{bodypart})} \\
\hline
 body in the eye   & 1258490 \\
 the eye           & 1175559 \\
 in the eye        &  852800 \\
 st aid : foreign  &  671132 \\
 body into the eye &  644635 \\
 aid : foreign     &  577192 \\
 his eye           &  477763 \\
 into eye          &  447310 \\
 in eye            &  421186 \\
 object in the eye &  419239 \\
\hline
\end{tabular}
}
\end{minipage}
\captionsetup{size=footnotesize}
\caption[Predictive regions example]{Top 10 predictive regions for the access and rules categories of  \texttt{incident\_type} (left and middle) and for the eye category of \texttt{bodypart} (right). Examples of valid precursors are in \textbf{bold}. \label{table:pred_regs}}
\end{table}

\noindent We can observe in Table \ref{table:pred_regs} that some of the predictive regions describe environmental or behavioral conditions that are observable \textit{before} incident occurrence, i.e., that are \textit{precursors} to incidents. We can see such examples in \textbf{bold}, for instance: ``trip hazard'', ``blocked by'', ``speeding in car park'',  ``slippery floor'', or ``not wearing''. This is of interest to the construction safety community as this approach shows a way how to use a trained model to \textbf{automatically learn valid \textit{incident precursors} from raw text}. These can be used to observe empirically what precursors are occurring and what incident/injury type they are related to.

Note that some of the regions in Table \ref{table:pred_regs}, such as ``access'' or ``his eye'', correspond to parts of the input text that explicitly mention the category to be predicted. Since the persons who wrote the incident reports often included the solutions (i.e. the outcomes) in the narrative, the model is, of course, using this information for prediction. Strategies to maximize identification of valid precursors, rather than outcome information, are outlined in subsection 7.1.\\

\noindent \textbf{Word saliency}. Another way to learn which parts of the input are predictive was first described by \cite{simonyan2013deep} for computer vision and applied to NLP by \cite{li2015visualizing}. The idea is to rank the elements of the input document $A \in \mathbb{R}^{s \times d}$ based on their influence on the prediction. An approximation can be given by the magnitudes of the first-order partial derivatives of the output of the model $\mathrm{CNN}: A \mapsto \mathrm{CNN}(A)$ with respect to each row $a$ of the input $A$:

\begin{equation}
\mathrm{saliency}(a)=\abs{\frac{\partial(\mathrm{CNN})}{\partial a}\mid_a}
\end{equation}

\noindent The interpretation is that we identify which words in $A$ need to be \textit{changed the least to change the class score the most}. The derivatives can be obtained by performing a single back-propagation pass based on the prediction, not on the loss like during training.

\begin{figure}[h]
\centering
  \fbox{\includegraphics[width=.85\textwidth]{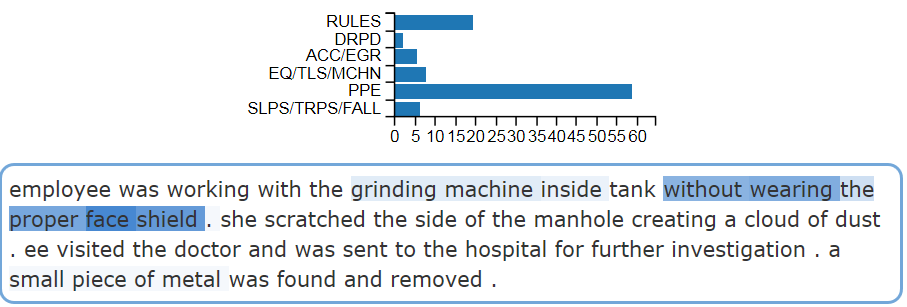}}
  \captionsetup{size=footnotesize,justification=raggedright}
  \captionof{figure}{CNN probabilistic forecast and word saliency visualization for \texttt{incident\_type}. \label{fig:saliency}}
\end{figure}

A visual example of this method is shown in Fig. \ref{fig:saliency} for the \texttt{incident\_type} outcome, where the darker highlight indicates the most predictive words, here ``without wearing'' and ``face shield''. Also, a probabilistic forecast is shown above the text. It corresponds to the output of the softmax layer.

To identify the most salient words at the corpus level, we used the same procedure as that described for the predictive regions method and recorded the most salient words for each report in the training set. In addition to single salient words, we also recorded salient phrases by pasting together the words that were following each other in the input text and which had much larger saliencies than the other words in the document.

\begin{table}[h]
\centering
\scalebox{0.75}{
\begin{tabular}{lr}
\multicolumn{2}{c}{Salient word sequences for FOB (\texttt{injury\_type})} \\
\hline
OOV : foreign body into the eye foreign body into the eye OOV OOV       & 142882 \\
st aid : foreign body . foreign body in the eye .                       & 121903 \\
something enter into eye OOV                                            &  61122 \\
st aid : foreign body in the eye 1 st aid : foreign body in the eye OOV &  52105 \\
went on his eye while \textbf{grinding} a                                        &  33956 \\
st aid : foreign body foreign body in the eye OOV                       &  22724 \\
OOV trauma into eye \textbf{pipe} fitter while                                   &  18296 \\
\textbf{sand} in eye worker ,                                                    &  17880 \\
foreign body in his right eye OOV OOV                                   &  17407 \\
: \textbf{sand} in eye 1 st aid : \textbf{sand} in eye                                    &  15768 \\
\hline
\end{tabular}
}
\captionsetup{size=footnotesize}
\caption[Word saliency examples]{Top 10 most salient word sequences for the FOB category of \texttt{injury\_type}. Examples of valid precursors are in \textbf{bold}. OOV denote out-of-vocabulary words. \label{table:sal1}}
\end{table}

\begin{table}[h]
\centering
\scalebox{0.75}{
\begin{tabular}{lr}
\multicolumn{2}{c}{Salient word sequences for Access (\texttt{incident\_type})} \\
\hline
 \textbf{trip hazards} on main                       & 2670 \\
 access to                                  & 2511 \\
 \textbf{slippery floor} deck 3                      & 2210 \\
 \textbf{slipping hazard} on stairs an               & 2061 \\
 a potential \textbf{trip fall hazard} deck          & 2050 \\
 the access to                              & 1933 \\
 safe access egress from main deck          & 1903 \\
 access \textbf{restricted access} to muster station & 1865 \\
 causing an access egress hazard in the     & 1843 \\
 a \textbf{trip hazard} on an                        & 1761 \\
\hline
\end{tabular}
}
\captionsetup{size=footnotesize}
\caption[Word saliency examples]{Top 10 most salient word sequences for the Access category of \texttt{incident\_type}. Examples of valid precursors are in \textbf{bold}. \label{table:sal2}}
\end{table}

High saliency words and phrases at the corpus level are shown in Tables \ref{table:sal1} and \ref{table:sal2}, where valid precursors are once again indicated in bold. By comparing those in Table \ref{table:sal2} for Access (\texttt{incident\_type}) to the ones found previously in Table \ref{table:pred_regs}, we can observe that while the predictive region and word saliency methods identify some common precursors, such as ``trip hazard'', they mostly extract complementary, rather than redundant, knowledge. For instance, Table \ref{table:pred_regs} identifies ``blocked by'' and ``obstructed by'' while Table \ref{table:sal2} contains ``slippery floor'' and ``slipping hazard''. Thus, we advise in practice to use both methods, rather than one or the other.

\subsection{HAN: word and sentence attention}
\begin{figure}[h]
\centering
\fbox{\includegraphics[width=0.85\textwidth]{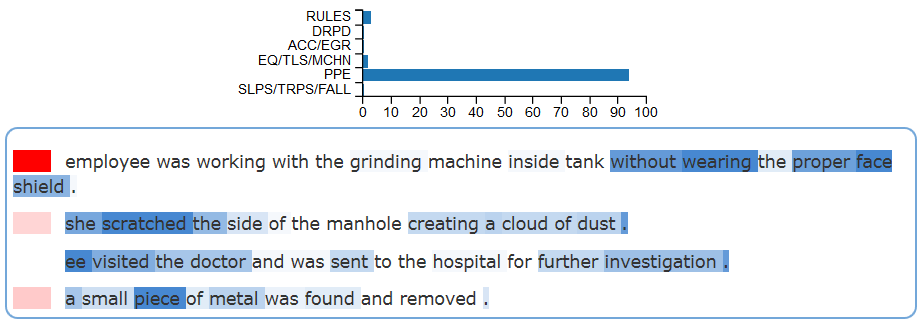}}
\captionsetup{size=footnotesize}
\caption{HAN probabilistic forecast, sentence and word attention, for the example of Fig. \ref{fig:saliency}. \label{fig:att}}
\end{figure}

For HAN, we can directly use the built-in word-level and sentence-level self-attention mechanisms to visualize the important words for a given report. An example is shown in Fig. \ref{fig:att}, where more important sentences are indicated by darker shading along the left side and more important words are highlighted darker within the sentence. In this way, we can identify the most important words in this example as ``without wearing proper face shield'' and ``piece of metal'' (dark sentence shading and dark word shading), which are also both valid precursors of the incident. 
To identify the most predictive words at the corpus level, for each report, we recorded the attention coefficient of each word multiplied by the attention coefficient of the sentence it belonged to. The scores were thus added across sentences and reports to gather the final statistics. Some examples are shown in Table \ref{table:att}.

\vspace{-0.3cm}

\begin{table}[h]
\centering
\begin{minipage}{0.28\textwidth}
\centering
\scalebox{0.75}{
\begin{tabular}{lr}
\multicolumn{2}{c}{PPE (\texttt{incident\_type})} \\
\hline
 eye            & 1066690 \\
 eye injury     & 750282           \\
 using          & 745095           \\
 worn           & 655728           \\
 ear protection & 645130           \\
 seat belt      & 619398           \\
 hard hat       & 612033           \\
 \textbf{sandblasting}   & 559528           \\
 lanyard        & 525512           \\
 quayside       & 513523           \\
\hline
\end{tabular}
}
\end{minipage}
\begin{minipage}{0.38\textwidth}
\centering
\scalebox{0.75}{
\begin{tabular}{lr}
\multicolumn{2}{c}{slips/trips/falls (\texttt{incident\_type})} \\\hline
 \textbf{water}               &      1816970 \\
 there               &      1539670 \\
 2017 - ankle sprain &      1278520 \\
 on                  &      1207330 \\
 low risk          &      1135350 \\
 \textbf{loose}               &      1102110 \\
 \textbf{step}                &      1094930 \\
 fall                &      1064280 \\
 \textbf{stairs}              &      1004320 \\
 \textbf{hole}                & 978237           \\
\hline
\end{tabular}
}
\end{minipage}
\begin{minipage}{0.3\textwidth}
\centering
\scalebox{0.75}{
\begin{tabular}{lr}
\multicolumn{2}{c}{FOB (\texttt{injury\_type})} \\
\hline
 \textbf{welder}              &      1596490 \\
 was                 &      1300760 \\
 at                  &      1178150 \\
 \textbf{dust}                &      1075000   \\
 ip                  & 949697           \\
 b \textbf{metal} went on his & 908895           \\
 ogp - fb r          & 789381           \\
 ips                 & 656907           \\
 cotton stick        & 639991           \\
 left                & 555208           \\
\hline
\end{tabular}
}
\end{minipage}
\captionsetup{size=footnotesize}
\caption[Word attention example]{Top words in terms of attentional coefficients at the corpus level. Examples of valid precursors are in \textbf{bold}. \label{table:att}}
\end{table}

\vspace{-0.2cm}

\subsection{TF-IDF+SVM}
Recall from subsection \ref{sub:svm} that the $\vec{w}$ vector, orthogonal to the best-separating hyperplane, is used to determine whether a given observation belongs to the class of interest or to any of the other classes (one-vs-rest approach). Here, observations are injury reports, and by definition, they can only have non-negative coordinates in the BoW space. Indeed, the minimum entry a report vector can have, for any dimension, is zero, when the corresponding token does not appear in the report. Thus, a given report belongs to the class of interest if its dot product with $\vec{w}$ is \textit{positive}.

Furthermore, the $\vec{w}$ vector contains the contribution of each token in making the classification decision. More precisely, the magnitudes of the coordinates of $\vec{w}$ indicate the strength of the contributions, while their signs indicate if the tokens attract or reject the report to/from the class of interest. Tokens for which the coefficients of $\vec{w}$ are large and positive (resp. negative) are strongly indicative of belonging (resp. not belonging) to the category of interest.

For each outcome, we recorded the most predictive $n$-grams. We made two interesting observations when analyzing the results: (1) the top of the lists of predictive tokens tend to contain $n$-grams of rather low order (mostly unigrams, then bigrams, and only rarely, trigrams); (2) more than for the deep learning approaches, one has to go down the list to start finding valid precursors, as the top 10 elements almost entirely relate to outcomes. In a way, it is like if the TF-IDF+SVM approach was more efficient in finding and focusing on the parts of the input text  containing the ``solutions'' than CNN and HAN. This could partly explain the observed superiority of TF-IDF+SVM on our datasets. 

However, when -slightly- going down the list, numerous valid precursors can be found. Examples are given in Table \ref{table:svm_featimp}. These examples were selected manually by inspecting the top 50 elements of each list.

\begin{table}[h]
\centering
\begin{minipage}{0.24\textwidth}
\centering
\scalebox{0.75}{
\begin{tabular}{lr}
\multicolumn{2}{c}{Rules {\footnotesize(\texttt{incident\_type})}} \\
\hline
smoking        & 3.822 \\
permit         & 3.759 \\
waste          & 2.803 \\
cigarette      & 2.591 \\
rules          & 2.576 \\
barricade      & 2.358 \\
speeding       & 2.254 \\
disposed       & 1.755 \\
chemicals      & 1.707 \\
rubbish        & 1.658 \\
\hline
\end{tabular}
}
\end{minipage}
\begin{minipage}{0.24\textwidth}
\centering
\scalebox{0.75}{
\begin{tabular}{lr}
\multicolumn{2}{c}{FOB {\footnotesize(\texttt{injury\_type})}}\\
\hline
dust            & 2.457 \\
splinter        & 2.238 \\
foreign body    & 1.789 \\
particle        & 1.605 \\
debris          & 1.12  \\
grinding        & 1.09  \\
leg steel pipe  & 0.955 \\
- insect    & 0.81  \\
irritated  & 0.789 \\
 visor  & 0.761 \\
\hline
\end{tabular}
}
\end{minipage}
\begin{minipage}{0.24\textwidth}
\centering
\scalebox{0.75}{
\begin{tabular}{lr}
\multicolumn{2}{c}{Eye {\footnotesize(\texttt{bodypart})}} \\
\hline
 dust           & 2.265 \\
 foreign        & 1.517 \\
 particle       & 1.373 \\
 flash          & 1.265 \\
 chemical       & 1.116 \\
 arc            & 1.009 \\
 leg steel pipe & 0.975 \\
 safety glasses & 0.947 \\
 welding flash  & 0.84  \\
 paint          & 0.831 \\
\hline
\end{tabular}
}
\end{minipage}
\begin{minipage}{0.24\textwidth}
\centering
\scalebox{0.75}{
\begin{tabular}{lr}
\multicolumn{2}{c}{Slips {\footnotesize(\texttt{incident\_type})}} \\
\hline
step            & 2.722 \\
stairs          & 2.487 \\
walkway         & 2.368 \\
oil             & 2.197 \\
grease          & 2.119 \\
spillage        & 1.927 \\
fluid           & 1.916 \\
 hole            & 1.9   \\
carpet          & 1.806 \\
handrail        & 1.796 \\
\hline
\end{tabular}
}
\end{minipage}
\captionsetup{size=footnotesize}
\caption{Hand-picked examples from the SVM top 50 most predictive $n$-grams. Most of them are valid precursors.\label{table:svm_featimp}}
\end{table}

\section{Limitations and recommendations}

\subsection{Recommendations for improving precursor extraction}
A current limitation of our methods, demonstrated in the results, is that models tend to use parts of the narratives explicitly discussing outcomes rather than the incident circumstances and environmental conditions. That is, the models have a tendency to ``cheat'' and use the solutions, which reduces their ability to extract meaningful precursors. Some remediation strategies, listed below, could be used:

\begin{itemize}
\item Using less text fields. In our experiments, we used four fields: \texttt{title}, \texttt{description}, \texttt{details}, and \texttt{root\_cause}. Information about outcomes can be found in all of them. However, on average, some fields, such as \texttt{title}, appear to contain more outcome information. Ignoring such fields could force the models to focus on information pertaining to environmental conditions and causes, and learn a higher proportion of precursors.
\item Removing from the input reports the text fragments that the models have most relied on the first time they were trained could be another solution. It would require, however, retraining models from scratch after the filtering step has been performed. This strategy would allow keeping more data than the first one, as it does not involve purely and simply ignoring fields. The drawback is that legitimate precursors might be removed along with the solutions during the filtering step. One would have to closely monitor the ratio $\nicefrac{\mathrm{precursors}}{\mathrm{solutions}}$ between the first and the second runs.
\item The last approach is very heuristic, but could work. It would consist in removing the parts of the reports following some keywords known to be indicative of outcomes (``hospital'', ``taken to'', ``injured person'', etc.). Alternatively, only the sentences containing such keywords could be eliminated.
\end{itemize}

\noindent There is a tradeoff between prediction performance and the usefulness of the features learned by the models. Giving the models access to text that may contain some elements disclosing information about the outcomes increases predictive performance, but the tradeoff is that the features learned by the models may be less useful from a knowledge extraction's standpoint. Conversely, implementing the strategies described above might result in a predictive performance drop, but the text fragments extracted by the models will include a higher number of valid precursors. Also, this setting is closer to a real-life prediction setting in which only information \textit{preceding} accident occurrence is available to the models.

Finally, note that while our models are inherently limited to extracting precursors present in the input documents, they are not limited to using short incident reports as input.
Longer, more detailed documents (including, e.g., root cause analysis or extensive description of the work conditions) can be used. Such documents are commonly available for medium severity (and more severe) reports, and could improve the quality and relevance of the precursors extracted.

\subsection{Recommendations related to prediction}
Predictive skill was not the focus of this study. We note, however, that combining the probabilistic forecasts of all models, for instance with a simple logistic regression, could generate more accurate predictions. This approach is known as \textit{model stacking}. 

Again, prediction was not the focus of this study. But if one was to use the models for prediction in the field, a user would have to write a textual description of the work conditions. This may be a cumbersome task that workers may not enjoy. Further, a computer, smartphone, or tablet with internet connectivity would be required. Such resources are not always available. However, with increasing computer literacy, availability of portable devices, and enhanced connectivity even in remote areas, these practical constraints may quickly resolve.

\section{Contributions to practice}
The methods proposed in this paper represent a substantial advancement in data-driven construction safety. They reduce the efforts needed to extract useful knowledge and make reliable safety predictions from injury reports. This information not only allows empirical relationships to be explored for post-analysis and project statistics, but such information can be used proactively during typical work planning, job hazard analyses, pre-job meetings, and audits. It also allows professionals to focus their safety efforts (and budget) towards reduction of certain incident types for specific activities by providing data to complement their expert knowledge. Given that injury reports are a ubiquitous form of data collected in industry, this advancement has immediate and significant impacts. 

The positive quantitative and qualitative results we observe despite the diversity of our dataset suggest that our approach is quite robust to using reports of various styles as input, including machine translated ones.
Thus, language and geographical differences are not a barrier to the applicability and transferability of our methodology.

Some of the automatically extracted precursors shown in section \ref{subsec:attimp} are trivial and thus already known to safety professionals.
We provided them mainly as a sanity check, to validate the approach.
Our approach is most useful when retrieving those less frequent, less obvious precursors, to complement the common knowledge ones.
Our approach is also very helpful when safety knowledge is scattered across the organization, due to the work being fragmented over projects, division, and geographical areas.
By processing the full database at once, our approach can put together a global, standardized view of safety issues, and make it readily available to all safety and management professionals, thus greatly improving communication and alignment.

Further developments could include using models to amalgamate reports across the industry, through continuous updates as new data arrive. This could allow the construction industry to collect learning without the need to access individual reports, often protected by confidentiality. Trends could be observed across companies and geographical boundaries. 

Also, the methods could be applied to other textual data, within or outside of construction safety. For example, contractual information or other data containing information of historical events such as lessons learned or quality records.

\section{Conclusion}
We proposed an approach to automatically extract valid accident precursors from a dataset of raw construction injury reports. Such information is highly valuable, as it can be used to better understand, predict, and prevent injury occurrence.
For each of three supervised models (two of which being deep learning-based), we provided a methodology to identify (after training) the textual patterns that are, on average, the most predictive of each safety outcome. We verified that the learned precursors are valid and made several suggestions to improve the results.
The proposed methods can also be used by the user to visualize and understand the models' predictions.
Incidentally, while predictive skill is high for all models, we make the interesting observation that the simple TF-IDF + SVM approach is on par with (or outperforms) deep learning most of the time.

\section{Acknowledgements} 
We thank the anonymous reviewers for their helpful feedback.

\section{Author Contributions} 
Antoine Tixier conceptualized and designed the study, curated the dataset, implemented the models, ran the experiments, interpreted and visualized the results, and wrote the core of the paper.
Henrietta Baker wrote the first part of the introduction, the literature review, the contributions to practice section, and the conclusion; reviewed the paper, and led the paper submission process. Matthew Hallowell reviewed the paper. 



\section{References}
\footnotesize
\bibliographystyle{elsarticle-num} 
\bibliography{bib}

\end{document}